\documentclass{svproc}
\usepackage{graphicx}
\usepackage{wrapfig}
\usepackage{url}

\newcommand{\rul}[1]{{\tt #1}}

\begin{document}
\mainmatter              % start of a contribution
\title{Don't Forget Imagination!}
\titlerunning{Don't Forget Imagination!}  % abbreviated title (for running head)
%                                     also used for the TOC unless
%                                     \toctitle is used
%
\author{Evgenii E.~Vityaev\inst{1} \and Andrei Mantsivoda\inst{2}}

\authorrunning{Evgenii E.~Vityaev, Andrei Mantsivoda} % abbreviated author list (for running head)

\institute{Sobolev Institute of mathematics SD RAS, Novosibirsk, Russian Federation,\\
\email{vityaev@math.nsc.ru},\\ WWW home page:
\texttt{http://old.math.nsc.ru/AP/ScientificDiscovery/}
\and
Irkutsk State University 1, K.Marx Str., Irkutsk 664003, Russian Federation \\
\email{andrei@baikal.ru}}

\maketitle              % typeset the title of the contribution

\begin{abstract}
Cognitive imagination is a type of imagination that plays a key role in human thinking. It is not a ``picture-in-the-head'' imagination. It is a faculty to mentally visualize coherent and holistic systems of concepts and causal links that serve as semantic contexts for reasoning, decision making and prediction. Our position is that the role of cognitive imagination is still greatly underestimated, and this creates numerous problems and diminishes the current capabilities of AI. For instance, when reasoning, humans rely on imaginary contexts to retrieve background info. They also constantly return to the context for semantic verification that their reasoning is still reasonable. Thus, reasoning without imagination is blind. This paper is a call for greater attention to cognitive imagination as the next promising breakthrough in artificial intelligence. As an instrument for simulating cognitive imagination, we propose semantic models -- a new approach to mathematical models that can learn, like neural networks, and are based on probabilistic causal relationships. Semantic models can simulate cognitive imagination because they ensure the consistency of imaginary contexts and implement a glass-box approach that allows the context to be manipulated as a holistic and coherent system of interrelated facts glued together with causal relations.
% We would like to encourage you to list your keywords within
% the abstract section using the \keywords{...} command.
\end{abstract}

\begin{flushright}
\textit{Imagination is more important than knowledge. \\ For knowledge is limited, whereas imagination \\embraces the entire world...\\
%, \\ stimulating progress, giving birth to evolution.\\
\textbf{Albert Einstein}}
\end{flushright}

\section{Introduction}

Human thinking is a complex mechanism that combines many faculties that effectively work together. AI technologies are gradually advancing in modeling these abilities. Phenomenal results have been obtained for problems formulated as auto-regression on discrete dictionaries. A surprisingly wide range of problems can be solved using this technology, including linguistic problems, various generation problems, solving mathematical tasks, programming, and, in general, problems of modeling human intuition. Things are moving more slowly in other areas, like modeling the human ability to navigate in a continuous omnidirectional real world, although great efforts are being made here too \cite{lecun}.

In this context, it looks very strange that so little attention is paid to such a crucial human faculty as \textbf{cognitive imagination}. 
% The general understanding of imagination in AI is rather ambiguous \cite{wiki}. 
Cognitive imagination \cite{cogim} is type of imagination that not a ``picture-in-the-head' but a faculty that provides coherent and holistic systems of concepts and causal links, that is, systems that serve as contexts for thinking, reasoning, decision making and prediction. Cognitive imagination is an ancient concept, the first version of which was proposed by Aristotle in \textit{De Anima}, who defined it as a mediator between sense perception and judgment that necessary for logic and reason.

Obviously, cognitive imagination is a completely different human ability than what, say, LLMs can do. LLMs are kind of a black box that pushes out answers on request, whereas imagination allows us to see whole pictures, either real or fictional. Loosely speaking, imagination provides a glass box, in which the semantic concepts of truth and false work. Within imagination, a person structures his or her imaginary pictures, identifies various concepts, object properties and connections between entities. The lack of interest in imagination looks even more strange if to think of all benefits  that  imagination models can bring to solving current AI problems (see section \ref{benefits}).

So, this paper is a call for greater attention to cognitive imagination, as we believe that efforts in this direction may lead to the next breakthrough in artificial intelligence. Below, we consider conceptual aspects and outline some ideas for modeling cognitive imagination.

\section{Gognitive imagination}
Imagination is the act or power of forming a mental picture of something not present. In terms of two systems of thinking coined by Daniel Kahneman \cite{kahneman}, imagination provides semantic context for deliberate thinking (as part of System 2), while itself is heavily dependent on intuition and previous experience (as part of System 1).

\begin{remark}
Just to recall, Kahneman considers human thinking as the interaction of two modes. System 1 is intuitive, fast, and operating with almost no effort. This mode allows us to quickly come to decisions based on our previous experience and taught patterns. In contrast, System 2 is conscious, deliberate, slow and takes a lot of effort. This mode is for the situations when we are faced with difficult problems or something unknown that we need to understand better. System 2 is also very useful for estimation of solutions generated by frivolous and swift System 1. Large language models have hugely boosted the simulation of human intuition, a black box that promptly produces solutions to our requests based on prior learning.
\end{remark}

\begin{wrapfigure}{r}{0.35\textwidth} %this figure will be at the right
    \centering
    \includegraphics[width=0.35\textwidth]{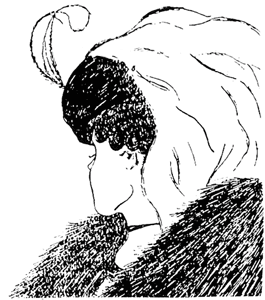}
    \caption{A young girl or an old woman?}
\end{wrapfigure}

Within System 2, imagination plays the crucial role in human deliberate thinking and reasoning: 
\begin{itemize}
    \item Imagination provides the semantic context of thinking: background info for reasoning together with tools for checking reasoning and easy switching from one context to another,
    \item Imagination supports the modeling of real and fictional worlds and their semantic structure,
    \item Imagination provides a holistic understanding of subject domains (``domains as a whole''),
    \item Imagination fosters creativity, offers a basis for innovative ideas and inventions,
    \item Imagination aids in visualizing scenarios and outcomes, helps in planning future actions,
    \item Imagination enhances learning, allows people to reveal connections between concepts and explore new ideas,
    \item Imagination allows individuals to better understand other individuals, their motivations and perspectives,
    \item Imagination offers a mental space for escaping from reality and safe fictional adventure.
\end{itemize}

The basic feature of contexts imagined by people is their consistency. A human's mind cannot simultaneously imagine contradictory things. This feature can be easily tested using the famous young-girl/old-woman illusion (Fig. 1). In this picture we can see a young girl or an old woman, but cannot see them simultaneously and must deliberately switch between the two to preserve consistency. The ability of switching between contexts/models is an incredibly important feature of the human mind \cite{vityaev5}. It allows us to split the vast world around us and our own fictional worlds into a plethora of contexts and focus our mind on solving specific tasks in specific domains when necessary.

Cognitive imagination is based on a semantic model of the world. When we imagine something, we see a picture from which we can estimate if some statement about this picture is true or false (or probable in more subtle cases). 

So, imagination plays a crucial role. Together with intuition and reasoning, it forms the triad of human consciousness, in which each of the components plays an irreplaceable role and closely interacts with the other two.

\section{Cognitive imagination and mathematics}
The next question is: what is a reliable environment for cognitive imagination modeling?

In his book \cite{tegmark}, Max Tegmark pronounced his belief that our Universe is actually a mathematical object. We are not ready to go as far as Tegmark, but believe in his other idea that a human, while speculating on the external reality, is based not on the reality itself but on the \textbf{internal reality}, that is, on an image that reflects the external reality in his or her mind. 

Tegmark's next major idea is a \textbf{consensus reality}, that is, a common piece of internal reality shared by a group of people. Consensus reality is the only instrument that allows us to understand each other and work together. Tegmark considers the real world, but this idea can be generalized to any domain, real or fictional. We call such generalizations \textbf{internal} and \textbf{consensus models}.

\begin{remark}
To illustrate consensus models, just imagine how the same facts might be interpreted by the audiences of Fox News and CNN. The facts are the same, the process of thinking is (probably) the same, but the results can be strikingly different. And the only difference here is the difference in consensus models. 
\end{remark}

As for the Tegmark's idea that the Universe is a mathematical object, we rather believe the opposite: mathematics is as it is because it catches something basic from the Universe we inhabit and something crucial in the way our minds work. For example, it is well known that a mathematical model cannot be inconsistent by definition (it cannot simultaneously make both \(F\) and \(\neg F\) true \cite{modeltheory}). And this entirely reflects the context consistency in our imagination \cite{vityaev4,vityaev5}. Actually, the different contexts in our mind can contradict each other (and, thus, pave the way for doublethink). But contradictions are unimaginable within a single context (remember the girl-woman illusion). To accept contradictions we need to switch between the contradictory concepts.

Interestingly enough, all this is deeply connected with reasoning. An individual always reasons in some context (i.e. a model), takes from the context the facts needed for reasoning, and finally returns to the context to semantically check whether he or she has gone astray in the process of reasoning. In other words, there is no reasoning in a human mind without imagination.

Note that the same idea is mathematically expressed in the G\H{o}del's completeness theorem \cite{modeltheory}: a set of statements is consistent (that is, \(F\wedge \neg F\) is not derivable from it) if and only if this set has a model. So, we can say that the G\H{o}del's theorem is a mathematical reflection of the abilities of our imagination. 
% So, mathematically imagination contexts can be established by semantic (logical) models. 

Actually, we believe that mathematics of semantic models (and perhaps mathematics as a whole) largely grew out of imagination. A human can imagine one context or another, mentally depict either the real world or imaginary worlds, and easily switch between them. In mathematics, models are also a key concept, and different axiomatic systems form different contexts and worlds. For us, this makes semantic models a priority tool for simulating cognitive imagination that manipulates concepts and links between them. 

Semantics is also closely related to imagination. To evaluate whether a statement is true or false, we simply check this statement against our internal picture. The same is true of reasoning - both the reasoning itself and its outcome depend entirely on the picture of the world that we have in our heads (remember CNN and Fox News). People borrow the ``axioms'' as starting points for reasoning from the contexts they keep in minds.

To finalize, a system of contexts and interpretations allows us to segment the huge external world and the internal world of our fantasies, to digest these worlds ``piece by piece'', as well as to semantically structure and conceptualize the vast and monotonous swamp of the reality. And mathematical semantic models can reflect this human faculty.

\section{What about GOFAI?}

Since the current reputation of mathematical methods as AI methods is problematic, before proposing yet another symbolic approach, it is better to understand why symbolic and logical approaches have not worked as expected in AI so far. Despite enormous efforts (e.g. the handbook on the resolution principle alone contains over two thousand pages \cite{resolution}), symbolic methods based on mathematical logic have failed to show results even slightly approaching those achieved with the help of neural networks. For this, they even got non-flattering description of ``Good old fashioned artificial intelligence (GOFAI)''.

In essence, we see three severe miscalculations here:

\begin{description}
    \item\textbf{The interpretation of logic inference as human reasoning is erroneous and misleading}. The reality is that the symbolic inference is the way of {\bf writing and verifying} results that have already been obtained. For instance, mathematicians heavily rely on their intuition while searching for a proof of statements, so System 1 has a strong influence on this process. Therefore for finding proofs, LLMs with their chain-of thought \cite{chain} can be more reliable (intuition manages to surpass infamous combinatorial big bangs in the search space). This is, probably, why hybrid approaches like GeometryAI \cite{gemai} perform better than purely symbolic approaches.
    \item[Two-valued logic is too rigid.] This is probably not the most crucial problem, because there are many (e.g. modal) logics with more sophisticated truth handling that are also unsuccessful in the context of AI. But in any case, true/false semantics is too inflexible to properly reflect real-life knowledge with fifty grades of uncertainty.
    \item[There is no learning.] This one is decisive. Classical models are insensitive, they lack reliable tools for learning, which is crucial for getting updated, feedbacked and better tuned to the ever-changing world.
    
\end{description}

We are sure that the basic problem of GOFAI is the wrong focus on a reasoning instead of a semantic modeling of imagination. The second problem is the lack of learning. GOFAI methods do not provide  mathematical models with the ability to learn. This is why the huge efforts to use traditional inference systems for solving reasoning tasks have largely failed.

\section{Semantic model architecture}

Thus, our position is that 
\begin{quote}
\begin{enumerate}
    \item Symbolic mathematics should not be used as a primary tool for reasoning, because it cannot use intuition (as part of System 1). 

    \item  But it can be great for \textbf{imagination modeling} (as part of System 2) because it allows one to see the domain as a whole and provides sound semantic evaluation and verification. 
\end{enumerate}
\end{quote}

So, below we propose the conception of a semantic model that simulates an imagination context and possesses the following features:
\begin{itemize}
    \item knowledge described in a semantic model is accessible as a whole in a {\bf glass box} style in contrast to the black box of neural networks;
    \item knowledge in a semantic model has sound and verifiable logical semantics: it allows the direct checking of statements in terms of truth and false;
    \item the knowledge in a semantic model is easily interpreted in natural language;
    \item a semantic model {\bf can be trained}.
    
\end{itemize}

\begin{remark}
Actually, an extremely simple version of such semantic models is now actively used in compound AI systems and agents. These are databases. All the ``knowledge'' in them is accessible and can be used as a whole, for instance, in analytical tasks. The semantics of databases is very primitive, but deterministic and verifiable. Via computations, we can check if this or that statement about them is true. And last but not least, they can ``be trained'' using imperative assignment, which brutally sets new values to properties. Of course, databases are too primitive to properly simulate imagination, in particular because their data do not provide a general tool for semantically seeing the integrated picture, so each time we have to develop domain-specific tools to achieve this.
\end{remark}

Primarily the ideas of semantic modeling were proposed in \cite{ges}. A state-of-the-art overview can be found here \cite{all,vityaev10}.

% In our journey we adhere to the following principles:
The architecture of semantic models is based on the following principles:

\begin{description}
    \item[Principle 1.] A \textbf{learning semantic model} (further, a semantic model for brevity) is a two-tier hybrid system, that reflects two types of domain knowledge:
    \begin{description}
        \item[Factual Model] that contains deterministic data/facts about the domain and reflects its basic parameters, objects and links between objects,
        \item[Causal Model] that contains causal relations as general knowledge about the domain; the level of certainty of causal relations can vary.
    \end{description}

    \item[Principle 2.] Factual Models are a kind of ``digital twins'' that reflect the current state of the domain, they evolve over time based on the perception of signals from external sources (``oracles'').

    \item[Principle 3.] Relations in the Causal Model are expressed in the formal language of the Factual Model or its conservative extension.

    \item[Principle 4.] Semantic machine learning consists of finding/updating causal relations  that reflect the current state and behavior of the Factual Model, and collecting them in the Causal Model.
    
\end{description}

The Factual Model is necessary to store specific and deterministic parameters about the current state of the domain. Factual knowledge provides a robust and structured framework for imaginary contexts, concepts and links between them. It evolves over time in response to external signals and internal processes.

The next question is why did we choose causal relations for representing the general implicit knowledge about the domain? It is well known that causal relations are very important for human thinking and, in particular, allow us to glue together different parts of mental pictures. Causality has a long history of study in philosophy, physiology and psychology starting with four types of causes of things in Aristotle's \textit{Metaphysics}. Much effort has been put into modeling causal relationships \cite{causal1}.  Causal relations are strongly connected to imagination \cite{cri,rehder,vityaev4,vityaev5}. We were curious to get a justification of this from an artificial mind, so asked ChatGPT to prepare the rationale for this. And here is what we got:

\begin{itemize}
    \item \textbf{Creating Scenarios}: Imagination allows individuals to construct scenarios based on causal relationships. This involves envisioning how one event can lead to another, enabling us to foresee potential outcomes. 
    \item \textbf{Problem Solving}: Imagination aids in solving problems by allowing individuals to mentally simulate different actions and their consequences. This helps in exploring various paths without real-world trial and error. (At this point, the authors recalled Karl Popper's thought that this kind of consciousness evolved so that our hypotheses would die instead of us.) 
    \item \textbf{Forecasting}: People use imaginative thinking to predict future events, relying on established causal connections. By imagining how current actions might impact future situations, they can make well-informed decisions. 
    \item \textbf{Storytelling}: Causal relations enhance narratives within imaginative stories. Writers create plots by linking events through causation, making stories relatable and compelling.
    \item \textbf{Learning and Adaptation}: Imagination fosters understanding by linking new information with existing knowledge through causal chains, facilitating deeper learning and adaptation to new experiences.
\end{itemize}

Great, everything is true. Causal relations play a crucial role in human thinking by helping individuals understand and predict events in their environment. And they do it while based on factual knowledge and previous exprience.

\section{Semantic machine learning}

The next obvious question is: how to make the concept of semantic modeling tangible? So, in this section we will briefly consider a hybrid mathematical formal system, which, from our point of view, can be used as an implementation of the general idea of semantic modeling. This solution is a hybrid system that integrates:
\begin{itemize}
    \item Object ontologies \cite{manpon1,manpon2} as factual models and
    \item Logic-probabilistic inference \cite{vityaev1,vityaev2} as a causal model.
\end{itemize}
This system, called semantic machine learning, was considered in \cite{gavman1}.

We propose semantic machine learning as the implementation scheme for semantic models, because it has a number of key properties. Thanks to probabilistic inference, it can deal with uncertainty. But, the fundamental problem of probabilistic inference is statistical ambiguity \cite{hempel}, when both \(F\) and \(\neg F\) can be inferred simultaneously.

In our works \cite{vityaev1,vityaev2} we investigated in detail, which probabilistic causal rules are the most accurate for predictions. The feature that distinguishes our approach from the others is that, as a result of these studies, we have solved the statistical ambiguity problem. There we defined maximally specific causal relations as a version of maximally specific statistical laws introduced by C.G. Hempel as the possible solution of this problem. In particular, we have proved that predictions based on these causal relations are consistent and cannot infer both \(F\) and \(\neg F\) simultaneously. In other approaches on causal relations (e.g. in [12-13]) the definition of causality does not guarantee the consistency of predictions. And it also plays an important role in the hybrid system that integrates probabilistic causal rule systems with ontologies.

The resulting semantic model naturally formalizes imagination \cite{vityaev3,rosch} as a logically consistent and prognostic model of reality \cite{vityaev3}, formalizes ``natural'' concepts \cite{rosch,rehder,vityaev4}, and implements G. Tononi's integrated information theory \cite{tononi}. This approach also allows for the automated discovery of new ``natural'' concepts \cite{redher2} that can semantically expand the ontology. 

Ontologies, in turn, are lightweight logic systems that provide basic knowledge about domains. Given a set of basic data types \(D\), an ontology provides a language over \(D\) that allows us to describe domain entities/objects, their basic properties (with values from \(D\)) and links between objects. Actually, object ontologies are more sophisticated in nature (see \cite{manpon1}), to make them suitable for building logical digital twins, but the basic features listed above are sufficient to illustrate our ideas.

As an example, consider the \textit{People and Hair} ontology, which describes several categories of objects (persons, haircuts, hair colors and people's occupations). The language of the ontology also contains properties 
\rul{HasHairColor(\(x\)) = \(y\)} (\(x\) has hair color \(y\)),
\rul{Age(\(x\)) = \(y\)} (\(x\)'s age is \(y\)) and
\rul{HasOccupation(\(x\)) = \(y\)} (\(x\) is occupied by \(y\)).
The ontology essentially consists of facts about people, their haircuts, ages, and occupation, including 
\begin{itemize}
    \item \textit{categorical facts} about objects such as \rul{Person(Victor)}, \\ \rul{HairColor(Brunette)}, \rul{Occupation(Musician)},
    \item \textit{object properties} such as \rul{Age(Ann) = 15}, where \rul{15} \(\in D\),
    \item \textit{links between objects} such as \rul{HasHairColor(Nataly) = \\ Blonde}, \rul{HasHairColor(Ann) = Green}, \\ \rul{HasOccupation(Nataly) = Researcher}.
\end{itemize}

These data are concrete, simple and deterministic, and the totality of facts determines the implicit general knowledge about the ontology and its inhabitants. Who are these people with green hair? What is known about the activities of children in the domain? Et cetera, et cetera. We must study and analyze the ontology to retrieve this knowledge. And here come \textit{causal relations} and \textit{semantic machine learning} on ontologies. 

\textit{Causal relations} in logic-probabilistic inference \cite{vityaev3} have the form 
\begin{equation}\label{first}
    R = T_1^\star(x) \;\land\ldots \land\; T_n^\star(x) \rightarrow T_0^\star(x)\;\;\;\;\;
\end{equation}
Here \(T_1, \ldots, T_n, T_0\) are \textit{concepts}, and $T_i^\star(x)$ is either $T_i(x)$ or $\neg T_i(x)$. Concepts comprise a language, in which we want to learn what is going on in the ontology. Formally, concepts are unary predicates that conservatively augment the language of the ontology. The relation \(R\) is probabilistic and \(p\) is its \textit{conditional probability}, which is defined as
\begin{equation}
  p(R) = p(T_0^\star | T_1^\star \;\land\ldots \land\; T_n^\star) = \frac{N(T_1^\star \;\land\ldots \land\; T_n^\star \land T_0^\star)}{N(T_1^\star \;\land\ldots \land\; T_n^\star)},
\end{equation}

\noindent
\(N(F)\) is the number of elements for which \(F\) is true. The conditional probability exists if $N(T_1^\star \;\land\ldots \land\; T_n^\star) > 0$.

Obviously, we are interested only in those causal relations \(R\), which have high conditional probability \(p\) and do not have redundant, superfluous conditions that also provide sufficient causal learning. Such relations are called \textit{probabilistic laws} \cite{vityaev1}. Loosely speaking, a probabilistic law for \(T_0^\star\) is an optimal rule (\ref{first}) with the maximal value of \(p\).
\begin{quote}
    \textbf{Semantic machine learning} is the process of finding probabilistic laws of an ontology.
\end{quote}

Let us get back to the \textit{People and Hair} ontology and introduce the following concepts:
\begin{itemize}
    \item[] \rul{\(T_1(x) \equiv\) HasHairColor(\(x\)) = Brunette}
    \item[] \rul{\(T_2(x) \equiv\) HasHairColor(\(x\)) = Blonde}
    \item[] \rul{\(T_3(x) \equiv\) HasHairColor(\(x\)) = Green}
    \item[] \rul{\(T_4(x) \equiv\) Age(\(x\)) \(< 16\)}
    \item[] \rul{\(T_5(x) \equiv\) Age(\(x\)) \(\geq 16\; \land\; \)Age(\(x\)) < 25} 
    \item[] \rul{\(T_6(x) \equiv\) Age(\(x\)) \(\geq 25\; \land\;\)Age(\(x\)) < 50}
    \item[] \rul{\(T_7(x) \equiv\) Age(\(x\)) \(\geq 50\)}
    \item[] \rul{\(T_8(x) \equiv\) Occupation(\(x\)) = Musician}
    \item[] \rul{\(T_9(x) \equiv\) Occupation(\(x\)) = Researcher}
    \item[] \rul{\(T_{10}(x) \equiv\) Occupation(\(x\)) = Student}
\end{itemize}
Here \(T_1(x)\) is the concept of being a brunette, \(T_4(x)\) is the concept of being very young, etc. Suppose, our ontology describes a group of people we are interested in learning about. Then semantic machine learning provides us with probabilistic laws such as

\begin{itemize}
    \item \(T_4(x) \rightarrow T_{10}(x)\;\;\; [[0.98]]\)
    \item \(T_3(x)\;\land\;T_5(x) \rightarrow \neg T_{9}(x)\;\;\; [[0.95]]\)
    \item \(T_7(x) \rightarrow \neg T_{3}(x)\;\;\; [[0.999]]\)
    \item etc.
\end{itemize}

The first law states that (in our particular ontology!) almost all children are students; the second ensures that there is almost no young researchers with green hair. 

Semantic machine learning is powerful enough to automatically create new meaningful concepts, as well as discover strategies and policies, explicitly formalize them as probabilistic rule systems, and support workflow control. 

Real-life problem solutions for causal relations of the form (2) were demonstrated in papers on financial forecasting, bioinformatics, medicine, and forensic accounting, see Scientific Discovery website \cite{SD}. 

\begin{wrapfigure}{r}{0.35\textwidth} %this figure will be at the right
    \centering
    \includegraphics[width=0.35\textwidth]{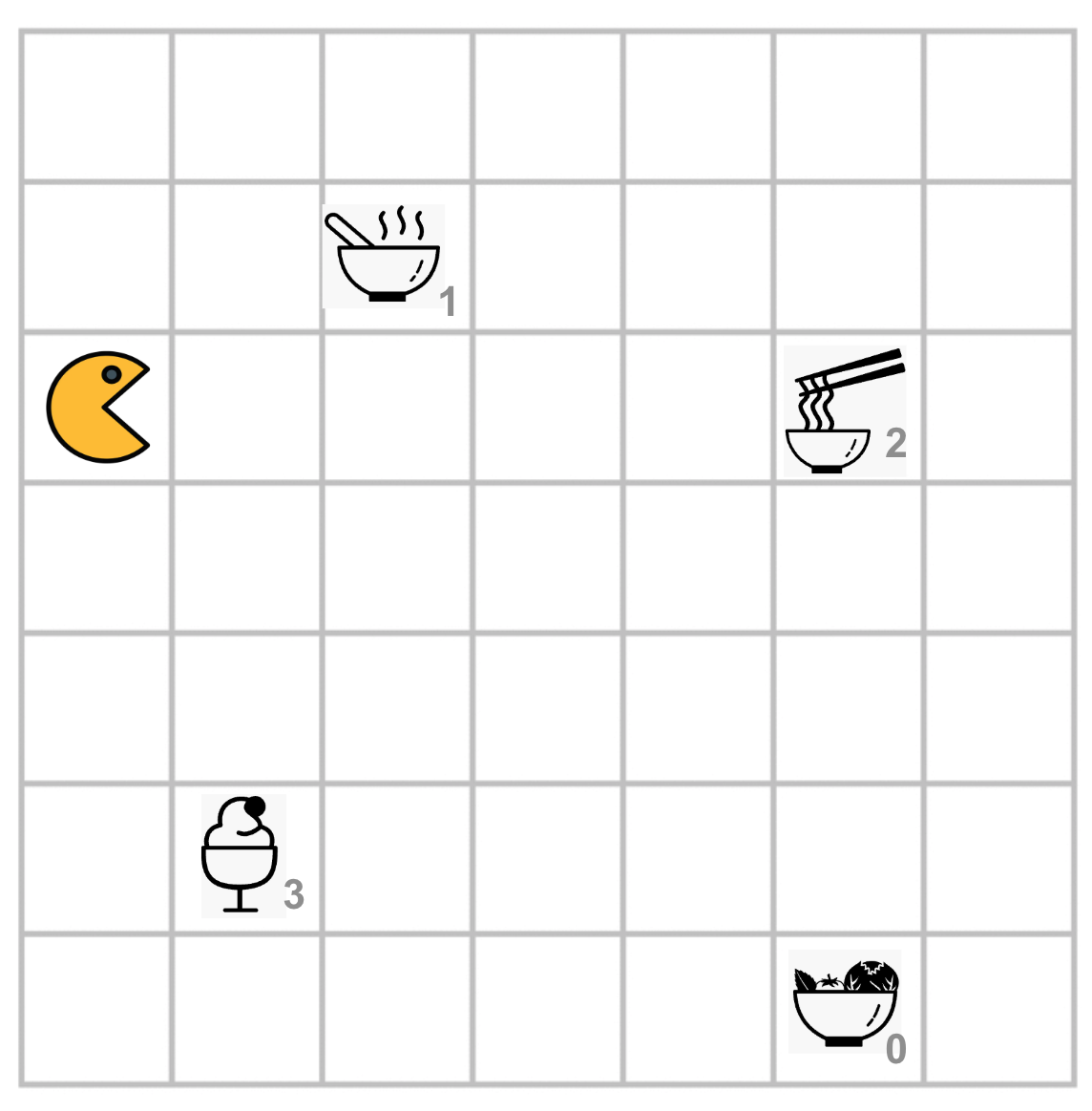}
    \caption{The lunch problem}
\end{wrapfigure}

To illustrate this, consider the following task that has an agent-like flavor \cite{dishes2}. Imagine a grid of cells of a fixed size and an ``agent'' that ``lives'' on this grid. At any given time, the agent occupies one of the cells. It can move forward, turn left and right. We want to train the agent to eat lunch in the same order as humans: starting with an appetizer and ending with a dessert, and eating a soup and a main course in between. The dishes are randomly distributed across the cells of the grid. The agent's task is to find a general strategy for lunch in situation when it does not have additional information, such as knowledge of the order of the dishes. To do this, it is necessary to invent subgoals and find causal relations between them (Fig. 2). 
When learning, the agent explores the grid and finds probabilistic rules such as 
\begin{equation}
Front(dessert) \wedge moveForward \rightarrow Center(dessert)  \wedge PickedUp
\end{equation}
\noindent
which states that if the agent sees the dessert in front of it, then it can move forward and eat it, that is, reach the final goal. Unfortunately, this rule has fairly low probability overall, because it requires the main course -- eating subgoal to be fulfilled. To increase this probability up to \(1.0\) semantic machine learning invents a new subgoal/concept \(Eaten(MainCourse)\) and the probabilistic law 
\begin{eqnarray}
Front(dessert) \wedge moveForward \wedge Eaten(mainCourse) \rightarrow \\  \nonumber Center(dessert) \wedge PickedUp
\end{eqnarray}
which, evidently, has conditional probability of 1. See \cite{dishes2} for more details.

Such agent-based semantic machine learning has significant properties:
\begin{itemize}
    \item when learning, the agent discovers an hierarchy of subgoals with no limitations on depth, together with corresponding causal relations,
    \item agentic learning enriches the ontology language,
    \item agentic learning does not use the reward functions, it only needs the main goal to achieve,
    \item agentic semantic models are inherently interpretable (see the next section).
\end{itemize}

\section{Benefits}\label{benefits}
Let us return to the general conception of cognitive imagination modeling and consider the AI problems that this conception can tackle.

\begin{description}

\item[\textit{A Priori} knowledge.]
The prior knowledge problem is a problem of how to provide LLMs with information in addition to the training data. AI models are trained on static datasets and have a ``knowledge cutoff''. On the other hand, humans constantly develop, evolve and update their internal models of \textit{prior} knowledge. Another ability of humans is to easily switch from one context to another.  Each context has its own language, including categories, relations and properties. And posterior knowledge of one action becomes prior knowledge for the next action, and the cycle begins afresh.

Similarly, ontologies provide the system with the domain-specific language that is understandable to humans. The ontological and probabilistic knowledge tiers of semantic models speak the same language, a language built from terms and concepts of the underlying domain. So, the result of machine learning is also expressed in these terms. This is what makes semantic machine learning self-explanatory.
Incorporating semantically structured data into the system as prior knowledge also has a beneficial effect on traditional AI problems such as memory limitations, continuous learning, interpretability of learning results, and, of course, trust issues.

\item[Dealing with uncertainty.] For traditional AI models, dealing with uncertainty is a significant challenge. On the other hand, humans easily cope with uncertainty in situations with insufficient or even inconsistent information. 

Semantic modeling is based on uncertainty management. For instance, the manipulation of the reliability threshold, when we cut off probabilistic laws with lower conditional probability, supports risk management in decision making and the ability to take actions in uncertain circumstances. 

\item[Self-explanations and interpretability.] This feature ensures that the system can explain its reasoning and decisions in a way that humans can understand. A semantic model consists of simple ``facts'' about a domain and more general causal relations. Both are natural from a human perspective and can be directly explained, for example, by representing them in natural language. Reasoning based on the consecutive application of probabilistic laws also can be directly translated into natural language texts or program code.
Semantic modeling can also provide reverse engineering tools that allow the knowledge hidden in neural networks to be approximated and revealed.

\item[Memory and Context Limitations.] Even advanced traditional AI models struggle with persistent memory. Semantic models have direct and supervised means for long-term context memorizing for both fact-like data about objects, their properties and links, and more sophisticated causal knowledge. 

\item[Continuous Learning.] Semantic models can learn continuously and adapt to new situations as humans do. First, the evolving ontology of the semantic model can change in response to external signals and data. Second, only those parts of the semantic model  that are relevant to changing data are retrained. The semantic model remembers all knowledge, including the current state and historical data, just like humans do.

\item[Trust Issues.] In semantic models the notions of truth and falsity, as well as the levels of reliability are explicitly controlled. Its formal verification methods can estimate the level of trust, and the extent to which we can rely on decisions in circumstances of uncertainty. The transparency of semantic models is also supported by the creation of natural explanatory texts that are readable by humans. 

\end{description}

\section{Conclusion}

Finally, let us put our position in a nutshell:

\begin{description}
\item[Don't forget imagination!] Imagination is a crucial component of thinking. The lack of cognitive imagination models in artificial intelligence is the cause of many of its problems, such as issues with prior knowledge, dealing with uncertainty, trust issues, dangling reasoning, interpretability traps, memory and context limitations, continuous learning etc.
\item[A semantic model representation of imagination.] A combination of semantically structured ``facts'' with causal relations based on factual knowledge and the discovery of implicit intrinsic relationships represents a robust architecture for establishing cognitive imagination in artificial intelligence. 
\item[Reasoning without imagination is blind.] Imagination provides a solid background for reasoning. Reasoning always operates in some context, and people always use that context, first, to begin reasoning and then to repeatedly semantically check, whether they are right or wrong in their assumptions. 
\item \textbf{A hybrid of logic-probabilistic inference and object ontologies as an implementation}. We propose a hybrid formalism that integrates logical-probabilistic inference and object ontologies as a reliable way to implement semantic models of cognitive imagination. 
\end{description}

\medskip
\noindent\textbf{Acknowledgments}. Evgenii E.~Vityaev: The work is financially supported by State Assignment ``Logical calculus and Semantics, Model theory and Computability'' \# FWNF-2022-0011.

 Andrei Mantsivoda: The work is financially supported by the Foundation for Assistance to Small Innovative Enterprises, grant 408GS1CTS10-D5/101511.

\end{document}